# USING A CORPUS FOR TEACHING TURKISH MORPHOLOGY

Altay Güvenir and Kemal Oflazer

# BILKENT UNIVERSITY

# Department of Computer Engineering and Information Science

Technical Report BU-CEIS-94-23

cmp-lg/9503001   1 Mar 1995

This work was supported by the NATO Science for Stability Project Grant, TU-LANGUAGE.

# USING A CORPUS FOR TEACHING TURKISH MORPHOLOGY

H. Altay Güvenir and Kemal Oflazer
Department of Computer Engineering and Information Science,
Bilkent University, 06533 Ankara
{guvenir,ko}@cs.bilkent.edu.tr

**Abstract**

This paper reports on the preliminary phase of our ongoing research towards developing an intelligent tutoring environment for Turkish grammar. One of the components of this environment is a corpus search tool which, among other aspects of the language, will be used to present the learner sample sentences along with their morphological analyses. Following a brief introduction to the Turkish language and its morphology, the paper describes the morphological analysis and ambiguity resolution used to construct the corpus used in the search tool. Finally, implementation issues and details involving the user interface of the tool are discussed.

## 1 INTRODUCTION

Language instruction delivered on computers is not new. Computer-assisted language learning has been used in foreign language instruction for some time with varying success. One of the important characteristics of a successful computer assisted language learning system is its ability to give example sentences about the use of grammatical features of the language [12]. The aim of this research is to design and develop a corpus search tool as a part of an intelligent computer assisted Turkish language tutoring environment,[1] which will focus on issues Turkish morphology, syntax and semantics. The main purpose of the search tool is to enable the user to see actual usage of words along with the morphological structure of Turkish words. When completed, the CATT environment will comprise several components, e.g., an intelligent drill and practice tutor [3], an on-line dictionary, and a corpus search tool.

The corpus search tool searches and displays sentences, from a given corpus, that contain words satisfying the features selected by the learner. In order to do this, all the words in the corpus are passed through a morphological analyzer once. Since each word is analyzed independently of the other words in the sentence, its morphological analysis may lead to ambiguities. These ambiguities are resolved using another interactive tool called *xtag*. The following step is the extraction of all the morphological features of the words into a structure called feature indexes. The corpus search tool, using the tagged corpus and the feature indexes, lists all the sentences containing the words matching the features selected by the learner. The tool also displays the morphological analysis of a word selected by the user to show how that word matched the features set. The block diagram of the corpus search tool is given in Fig. 1.

---

[1]The computer-assisted Turkish tutor (CATT) is being implemented as a part of the TU-LANGUAGE project launched for developing computational foundations and applications for natural language processing in Turkish.

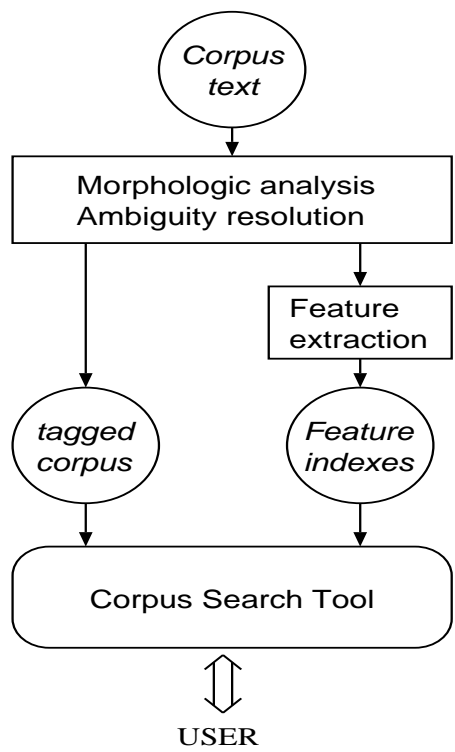

Figure 1: The block diagram.

The next section describes the corpus used in the teaching environment. The third section presents a brief introduction to the morphology of the Turkish language. Morphological analysis of words and the resolution of ambiguities are described in the fourth section. Section 4 presents the organization and the user interface of the corpus search tool. The final section concludes with our plans for future improvements on the tool.

## 2 CORPUS

For pedagogical issues, the corpus that will be used in a teaching environment needs to be selected with extra care. The sentences should be simple, yet descriptive, concise and instructive. The usage should be accepted by the experts on the language and its grammar. The sentences must not have any ambiguities.

Considering all the issues mentioned above, we decided to use, as the corpus for our tool, the example sentences describing the use of the entries in a dictionary for the Turkish language. The particular dictionary we chose is the "Türkçe Sözlük" published by Türk Dil Kurumu (Turkish Language Society) [2]. This dictionary contains about 20,000 exemplary sentences and currently, our corpus contains the first 1600 sentences of this dictionary for a total of 10984 words.

## 3 TURKISH MORPHOLOGY

Turkish is an agglutinative language with word structures formed by productive affixations of derivational and inflectional suffixes to root words. A simple example of a Turkish word formation is:

kesilemedi

which can be broken down into morphemes as follows:

| kes | +il | +eme | +di | |
|---|---|---|---|---|
| cut | +PASS | +NEG-CAP | +PAST | +3SG |
| stop | | | | |

This verb can be translated into English as "it could not be cut/stopped." For the details of Turkish grammar and word formations rules one can refer to a number of books [6, 13].

Turkish has finite-state but nevertheless rather complex morphotactics. Morphemes added to a root word or a stem can convert the word from a nominal to a verbal structure or vice-versa, or can create adverbial constructs. The surface realizations of morphological constructions are constrained and modified by a number of phonetic rules. Vowels in the affixed morpheme have to agree with the preceding vowel in certain aspects to achieve *vowel harmony*, although there are a small number of exceptions. Under certain circumstances, vowels in the roots and morphemes are deleted. Similarly, consonants in the roots words, or in the affixed morphemes undergo certain modifications, and may sometimes be deleted. The assimilation of a large number of words into the language from various foreign languages – most notably French, Arabic and Persian – have resulted in word formations which behave as exceptions to many rules. Turkish morphology has been investigated from a computational point of view by Hankamer [4], and by Solak and Oflazer [9, 10, 11], and by Oflazer [7].

## 4  MORPHOLOGICAL ANALYSIS AND AMBIGUITY RESOLUTION

Our morphological analysis subsystem is based on the two-level morphological specification of Turkish [7] based on the PC-KIMMO system [1]. This morphological analyzer is fairly comprehensive and is based on a root word lexicon of about 24,000 roots words. We can illustrate some of the phonetic phenomena of Turkish by a few examples:

| **Lexical:** | `masa+Hm` | `N(table)+1PS-POSS` |
|---|---|---|
| **Surface:** | `masa00m` | `masam` |

| **Lexical:** | `ev+HmHz+yA` | `N(house)+1PL-POSS+DAT` |
|---|---|---|
| **Surface:** | `ev0imiz+0e` | `evimize` |

| **Lexical:** | `ayak+nHn` | `N(foot)+GEN` |
|---|---|---|
| **Surface:** | `ayağ00ın` | `ayağın` |

Here we see in the first example that a high vowel at the beginning of a suffix (denoted by `H` representing {`ı`, `i`, `u`, `ü`}), drops when the stem it is being affixed to ends with a vowel. In the second example, we see that `H`'s are resolved as `i` in harmony with the last vowel in the stem `ev`, and the low, unrounded vowel (denoted by `A` in the last suffix, representing {`a`, `e`}) is resolved as `a` in harmony with the last vowel in the stem `evimiz` while the consonant `y` at the beginning of that suffix drops as the stem ends with a consonant. In the third example, we see that the last `k` at the end of a word becomes a `ğ` (a glide that forms bi-syllabic two-vowel sequences) when a morpheme starting with a vowel is affixed. There are numerous other such phenomena which make Turkish morphology rather complex to teach.

| nominal root | plural suffix | possessive suffix | case suffix | relative suffix |

Figure 2: The nominal paradigm

| verbal root | voice suffixes | negation suffix | modal suffix | main tense/aspect suffix |

| question suffix | second tense/aspect suffix | agreement suffix |

Figure 3: The verbal paradigm

The morphotactic component or the morphology is also reasonably complex. Turkish has two main paradigms for word formation. The *nominal paradigm* applies to nouns and adjectives, and derivations resulting in such parts-of-speech, and describes (loosely, as there are many exceptions and order variations) the order of the inflectional suffixes. This paradigm is described in Figure 2. The verbal paradigm applies to verbs and describes the order of the inflectional suffixes that are applicable to verbal stems. It is shown in Figure 3. The actual morphemes that are used and their sequence depend on the morpho-syntactic context.

Besides the phonetic and morphotactic phenomena, another difficulty that arises is the morphological structure and part-of-speech ambiguity. Typically Turkish words are on the average two-way ambiguous either with respect to their morphological structure or with respect to their part-of-speech, or both, for various reasons. In English, for example, a word such as *make* can be verb or a noun. In Turkish, even though there are ambiguities of such sort, the agglutinative nature of the language usually helps resolution of such ambiguities due to morphotactical restrictions. On the other hand, this very nature introduces another kind of ambiguity, where a lexical form can be morphologically interpreted in many ways. For example, the word *evin*, can be broken down as:

|    | **evin**        | **POS** | **English**    |
|----|-----------------|---------|----------------|
| 1. | N(ev)+2SG-POSS  | N       | (your) house   |
| 2. | N(ev)+GEN       | N       | of the house   |
| 3. | N(evin)         | N       | wheat germ     |

If, however, the local context is considered it may be possible to resolve the ambiguity as in:

|     | ... | senin | evin |
|-----|-----|-------|------|
|     |     | PN(you)+GEN | N(ev)+2SG-POSS |
|     |     | (your | house) |

|     | ... | evin | kapısı |
|-----|-----|------|--------|
|     |     | N(ev)+GEN | N(door)+3SG-POSS |
|     |     | ( door of | the house) |

We have developed a POS tagger for Turkish text based on our two-level specification of Turkish morphology augmented with a multi-word and idiomatic construct recognizer, and most importantly a morphological disambiguator based on local neighborhood constraints, heuristics and limited amount of statistical information [8]. The tagger also has additional functionality for statistics compilation and fine tuning of the morphological analyzer, such as logging erroneous morphological parses, commonly used roots, etc. Preliminary results indicate that the tagger can tag about 98% of the texts accurately with very minimal user intervention, by using about 200 usage and contextual constraints, and heuristics.

For morphological processing of our corpus we used our tagging tool *xtag* so that each word in the corpus had one and the correct morphological structure and part-of-speech. Morphological analysis and ambiguity resolution in our experimental corpus, which contains 10984 words in 1600 sentences, takes about 100 minutes of cpu time, as PC-KIMMO is a very slow morphological analyzer (about 2 words/second on Sun SparcStations)[2] The feature extraction on this corpus takes 7 minutes of cpu time.

## 5   THE CORPUS SEARCH TOOL

As we mentioned earlier, Turkish is an agglutinative language where words are formed by a sequence of morphemes that get attached to a root like "bead on a string." Hence, one of the more difficult issues in learning the grammar of the Turkish language as a foreign language, is the order of the morphemes in a word. The corpus search tool, called *corpus searcher* is designed to help the learner by displaying the order of the morphemes corresponding to a selected set of features. The words are shown as used in a sentence which is selected from the corpus. Illustrating the usage of words helps the learner to visualize the meaning of the whole sentence and the role of the morphemes in the words satisfying the features selected by himself.

The use of the corpus searcher is very simple. The user sets the values of some features of interest, then starts the search. All the sentences which contain a word satisfying the morphological features set by the learner are displayed in a window. [3]

The words matching the features are shown in bold font. If the learner clicks on any of these sentences, the morphological analysis of the matching word in that sentence is shown in a different window, along with its all morphological features.

---

[2]Currently, the speed of the tagger is limited by essentially that of the morphological analyzer, but we have ported the morphological analyzer to the XEROX TWOL system developed by Karttunen and Beesley [5]. This system which can analyze Turkish word forms at about 500 forms/sec on Sun SparcStation. We intend to integrate this to our tagger soon, improving its speed performance considerably.

[3]For the time being, the special characters in Turkish, namely, ç, ğ, ı, ö, ş, ü, are displayed using the nearest ASCII character in upper case.

As an example, suppose that the learner sets the value of the agreement feature to `3rd singular`, the aspect to `past`, and the voice to `passive`. Out of 1600 sentences in our current corpus 44 contain a word satisfying these conditions. Suppose further that the learner clicks on the sentence

> Musluğun akıntısı bir türlü **kesilemedi**.

which means "The leaking of the faucet could not be stopped, despite all efforts." The display of the corpus searcher would then ben as shown in Fig. 4. The word "kesilemedi" is shown in bold since it is the word that satisfies the morphological features that are set.

Currently, the corpus searcher enables the learner to specify any subset of the following set of features: agreement, aspect, case, category, possessive, sense, tense, voice, suffix, and root. The learner is free to set any subset, although the choice of some features imply the values of some others. For example, if the learner sets the value of case to be dative, the value of the category is implied to be noun. Therefore, any further selection on, for instance, the tense feature would result in the null set of sentences.

The program runs on Sun SparcStations. It is implemented in Lucid Common Lisp, with lispview (Xview) interface.

# 6   CONCLUSIONS AND FUTURE WORK

We have presented our initial implementation of a corpus search tool for teaching aspects of Turkish morphology. The system enables a learner to set certain morphological features and then searches a corpus of example sentences for words matching the specified features, and if necessary, also displays the morphological structure of the word. This system is planned to be a part of a much larger system, CATT, that is being designed for computer aided tutoring of the Turkish language.

In the near future, we intend to augment the morphological analysis display with an English translation of the matching word. This is in general rather non-trivial as the English translation for a single Turkish word may be a complete sentence. We also plan to enlarge the set of features and group them under a paradigm menu, and incorporate derivational suffixes so that the learner may also have a chance to see the semantic changes introduced by derivational suffixes. We also plan to have a much better user interaction model whereby the system gives clear messages if the learner sets the features in a conflicting manner that can not be realized in Turkish morphology.

# 7   ACKNOWLEDGMENT

This work is being done as a part of the TU-LANGUAGE project that is being supported by NATO Science for Stability program III.

Figure 4: The corpus searcher.